# Improving Link Prediction in Social Networks Using Local and Global Features: A Clustering-based Approach

S.Ghasemi[*], and A.Zarei[*]

[*] Department of Computer Engineering, Sepidan branch, Islamic Azad university, Sepidan, Iran.

**Abstract**

Link prediction problem has increasingly become prominent in many domains such as social network analyses, bioinformatics experiments**,** transportation networks, criminal investigations and so forth. A variety of techniques has been developed for link prediction problem, categorized into 1) similarity based approaches which study a set of features to extract similar nodes; 2) learning based approaches which extract patterns from the input data; 3) probabilistic statistical approaches which optimize a set of parameters to establish a model which can best compute formation probability. However, existing literatures lack approaches which utilize strength of each approach by integrating them to achieve a much more productive one. To tackle the link prediction problem, we propose an approach based on the combination of first and second group methods; the existing studied works use just one of these categories. Our two-phase developed method firstly determines new features related to the position and dynamic behavior of nodes, which enforce the approach more efficiency compared to approaches using mere measures. Then, a subspace clustering algorithm is applied to group social objects based on the computed similarity measures which differentiate the strength of clusters; basically, the usage of local and global indices and the clustering information plays an imperative role in our link prediction process. Some extensive experiments held on real datasets including Facebook, Brightkite and HepTh indicate good performances of our proposal method. Besides, we have experimentally verified our approach with some previous techniques in the area to prove the supremacy of ours.

## 1. Introduction

Social networks can be modeled as graph structures with nodes and edges to indicate the entities and their interactions, respectively. These interactions can be ideas, financial exchanges, friendships, and web links (Barabási, 1999). Two major services are applied in social networks; the first one is web based social networks which enables accessing via mobiles such as Facebook and Myspace. The second one is local services that operators of mobiles provide. By growing web and mobile usages, more members connect, share, and search for social networks. The structures of these networks enable the extraction of information to analyze the future behavior of members (Burt, 2013), however, it is to be noted that these structures are mostly noisy, distributed, unstructured and dynamic. The link prediction problem is related to finding unobservable links, missing links or near future interactions among members of a social network (Peng, BaoWen, YuRong, & XiaoYu, 2015), (Haghani & Keyvanpour, 2019) and (Bastami, Mahabadi, & Taghizadeh, 2019).
Recently, the link prediction issue is a topic that has been widely studied in social networks and a wide range of research has been developed, classified into three major categories including 1) similarity based approaches (Symeonidis, Tiakas, & Manolopoulos, 2010) which determine the proximity between any pair of neighbor nodes in a graph in order to extract similar nodes; 2) learning based approaches (Haghani & Keyvanpour, 2019), (Raut, Khandelwal, & Vyas, 2020) which present a model learned with a given feature and extract patterns; they eventually lead to link prediction and aim at abstracting the underlying structure of the input graph. Based on the type of learning model, these approaches are categorized into classification model and latent-feature-based model; 3) probabilistic statistical approaches (Samad, Qadir, Nawaz, Islam, & Aleem, 2020), on the base of probability and statistical analysis, optimize a set of parameters to establish a model which can best compute formation probability. In (Liben-Nowell & Kleinberg, 2007), significant improvements of performance of link prediction are investigated by the means of the timestamp of previous interactions. Backstrom proposes a service for peer-to-peer content sharing (Backstrom, 2007), which suggests friends with more knowledge about the contents; such research has valuable application in co-authorship fields. Another work (Zhou, Lü, & Zhang, 2009), has suggested a method to increase the efficiency and accuracy of the link prediction approach by assigning multiple topological criteria (based on time constraints); it applies five learning methods to training and experiment data. The results of the applied approaches indicate that learning methods improve the prediction accuracy up to 90%. On the other hand, Bliss (Bliss, 2014) uses a statistical approach for link prediction by periodically sampling the datasets and forming variance matrix and proximity matrix after extracting topological characteristics of nodes. Likewise, Zarco et al. (Zarco, Santos, & Cordón, 2019) apply visualization methods based on social network analysis techniques to get visual representations of the similarity relations of different companies on Twitter. It is worth mentioning that most of the existing similarity based works lose rich information available for link prediction by just considering the connections between the target nodes and their common neighbors. In addition, such lack of global similarity information in a graph kernel based link prediction method (Yuan, He, Guan, Zhou, & Li, 2019) exists to predict the sign of links through comparing user's similarity by using structural information of signed social network. Although a variety of works has been performed in this field, some major challenges in link prediction problems including nodes with low density links (Buchegger & Datta, 2009), scale of the network (Ilyas, 2008), and network dynamicity (Sharma, Datta, DeH'Amico, & Michiardi, 2011) already demand a load of endeavors.

Among existing methods, local similarity indices are the most popular ones that take into account the information of common neighbors to estimate the likelihood of existence of a connection between two nodes. The integration of local and global similarity measures can improve both the accuracy and the time complexity of link prediction methods (Bastami, Mahabadi, & Taghizadeh, 2019). The major disadvantage of current link prediction methods based on mere local indices is the low accuracy of predictions since these methods depend on the application domain (Yaghi, Faris, Aljarah, Ala'M, Heidari, & Mirjalili, 2020). Based on the fact that the behavior of the network is affected by its both structure and topology, recent studies have mostly focus on these characteristics of social networks. Topology of a social network which has a direct impact on the way information is transformed is governed by three factors: popularity of the nodes, their similarity and the attraction induced by local neighborhoods; structure refers to the way in which nodes that compose the network are interconnected. Some of the well-known structure based prediction methods are Common Neighbor, Jaccard's Index, Katz, and so forth. The dynamic characteristics of social networks have rarely been studied; e.g. lack of attention to density of networks causes unbalanced data and facing incorrect predictions, on the other hand, high scale of networks results in loss of time and accuracy due to the dispersion of information, or ignoring the dynamicity of the network loses relevant information of users' behaviors which leads to loss of accuracy. This paper extracts topological characteristics, such as local and global similarity measures, combines them with dynamic characteristics of nodes and applies them to learning based approaches in order to achieve an accurate and speedy link prediction process.

Link prediction has been mostly investigated through just one of the mentioned categories of methods, based on similarity, learning or probability. However, the link prediction process can also be tackled by using the combination of similarity based approaches (using an integration of local and global similarity metrics calculated for neighboring nodes) and classification based learning approaches. To strengthen the task of link prediction, our proposed two-phase approach in this research is based on both local and global similarity indices and classification methods as well. After preprocessing dataset, feature extraction is performed in first phase based on the topology characteristics and behavior of the network; the impact of each feature is measured to increase the accuracy of the prediction. Finally, in second phase, the link prediction process is applied using AdaBoost combined with decision tree. Since the existence of multi-dimensional attributes, including topological attributes, time attributes, and status attributes, a separate learner is applied to each cluster of attributes in this stage. To increase the speed, learners simultaneously process these clusters of features; ultimately, the clusters are combined to have an accurate link prediction. The innovations of our approach are 1) extracting the characteristics of the links to improve their prediction by introducing new similarity measures including local and global indices related to the position of nodes, dynamic behavior of the nodes and combining them with topological characteristics of the network in a way to increase the accuracy of the process, 2) applying learning based link prediction approaches to similarity based extracted measures. While previous researchers have considered some topological features and an image of the network in a specified moment for link prediction, we take into account the combination of the topological features of the network with the dynamic characteristics of nodes and clustering based on the most effective measures extracted in first phase of proposed link prediction approach. Our experiments have presented an efficient increasing of the accuracy of link prediction process.

The rest of this paper is organized as follows. The problem statement, link prediction concept, and previous researches are studied in Section2; Section3 studies the formulated problem of the link prediction, and the proposed approach. Section4 evaluates the efficiency of the solution. The paper is ended with some concluding remarks in Section5.

## 2. Link Prediction in Social Networks

### 2.1 Preliminaries

A social network is a social structure that consists of individuals or groups of entities interacting with each other based on one or more dependencies, such as ideas, financial exchanges, friendships, and web links. The structure is abstracted to a graph $G = (V, E)$, where $V$ is the set of vertices and $E$ is the set of edges as follows.

$e = (u,v) \qquad e \in E, u,v \in V$ (Eq. 1)

The communications are formed based on social interactions which can be direct or indirect. These relationships present interests of each user in interacting with other users.

In a social network structure, vertices are the individual and edges are the relations between these individuals in the network. Studying the probability of existence of a link between two nodes is a major challenge in social networks. For instance, in bioinformatics, there is very little information about inter-molecule transactions; instead of examining all possible interactions between them, it is possible to find a relatively accurate estimation of the link between the molecules using link prediction techniques. Social networks analysis deals with both individuals (people, organizations, states) as discrete analytical units of networks, and the way that the structure of disciplines affects the individuals and their relationships. Despite the analysis methods, some methods study the influence of the structure on the social norms. The structure of a social network determines the utility of the network for the individuals. Individuals with more relations to other social networks are more likely to have access to a wider range of information rather than having many connections within a network; this means that in social networks analysis, the properties of individuals have less importance than their communications. Link prediction is a general way to analyze social networks.

The scientific definition of link prediction is to seek for the added/removed edges to/from the network during time period $\Delta t$ (Liben-Nowell & Kleinberg, 2007). There are a load of works that study link prediction problem; a comprehensive review of link prediction approaches is provided in (Samad, Qadir, Nawaz, Islam, & Aleem, 2020) and (Mutlu & Oghaz, 2019). Researchers mostly categorize link prediction approaches into similarity based methods, learning based methods and probability statistical ones. Lu et al.

(Lü & Zhou, 2011), categorize link prediction into similarity-based, probabilistic and stochastic, algorithmic, and preprocessing methods.

### 2.1.1 Link Prediction on Social Networks

Apparently, many biological, social, and information systems can be described in form of networks, whose nodes represent individuals, and the links represent relations between individuals. Recently, a considerable amount of attentions have been devoted to the analysis of complex networks. Analyzing such networks, e.g. social networks, faces two challenges: incompletion (only part of social information can be gained from social networks) and dynamicity (some nodes and edges might appear or disappear in the future). Link prediction is the process of predicting the missing or unobserved links in current and future networks (Peng, BaoWen, YuRong, & XiaoYu, 2015), (Bastami, Mahabadi, & Taghizadeh, 2019).

Link prediction can be applied to many applications such as retrieval of missing information, identifying fake interactions, recommendation systems, network reconstruction, network classification, and network evolution mechanisms. The application of link prediction is much wider than the previously named domains; for instance, the proposed model, in (Almansoori, et al., 2012), is evaluated in two important interrelated applications of link prediction, namely health care and gene expression networks. Link prediction task considers both emerging and shrinking future links. Link prediction gene expression networks and medical referral (referring patients to physicians with specific area of specialty) are widely studied in (Almansoori, et al., 2012). Besides, link prediction has been exploited in bioinformatics in gene identification (Pittala, et al., 2020), gene expression (Turki & Wang, 2015), and drug response prediction (Stanfield, Coşkun, & Koyutürk, 2017). Furthermore, criminal network is a potential domain for link prediction process which analyzes network of criminal groups to identify some effective strategies to achieve network destabilization or disruption (Berlusconi, Calderoni, Parolini, Verani, & Piccardi, 2016). Moreover, the problem of estimating the robustness of link prediction algorithm in criminal networks is deeply studied by Calderoni et al. (Calderoni, Catanese, Meo, Ficara, & Fiumara, 2020). Additionally, in (Lim, Abdullah, Jhanjhi, Khan, & Supramaniam, 2019) criminal networks are formulated as a link prediction problem in order to outline the efficiency of temporal information on evolution of these networks in predicting future links by the aim of deep reinforcement learning techniques. The use of link prediction in authorship network of scientific publications (Liben-Nowell & Kleinberg, 2007), (Sharma & Minocha, 2016) and the citation impact (Klimek, Jovanovic, Egloff, & Schneider, 2016) can be mainly referred as other applications of this issue that predict the potential co-authors. Meanwhile, to have more productive recommendations in e-commerce, link prediction techniques can greatly benefit merchants and customers (Xie, Chen, Shang, Feng, & Li, 2015), (Li, Zhang, Meng, & Li, 2014). The application of link prediction on a wide range of social networks are assessed in (Yaghi, Faris, Aljarah, Ala'M, Heidari, & Mirjalili, 2020); these datasets include political blogs networks, biological networks, airport traffics networks, face-to-face contacts networks and co-authorship networks.

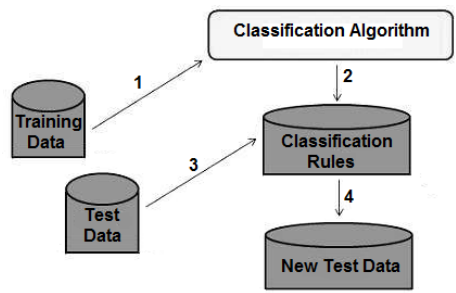

Fig. 1. The classification process

One of the most common ways to predict a link is to use a similarity based algorithms which use a simple similarity measure, and some new links would be predicted by ranking all the pairs of the nodes based on their similarity measures (Sharma & Minocha, 2016). The common description of similarity is the amount of relevant direct or indirect edges between nodes (Martínez, Berzal, & Cubero, 2017). The similarity-based link prediction approaches, on which we focused, are grounded in a measure of proximity or similarity between two entities, based on the network topology. Regarding the network being examined, computing the similarity indices might be very simple or very complicated (Lü & Zhou, 2011). Similarity means more common features (Liben-Nowell & Kleinberg, 2007), (Sharma & Minocha, 2016); however the features are usually hidden and must be extracted by using different techniques such as neural network concept (Sharma & Minocha, 2016) and Graph Neural Networks (GNNs) (Gaudelet, et al., 2020) known as powerful tools for learning hidden features in networks (Islam, Aridhi, & Smail-Tabbone, 2020). Likewise, Yaghi et al. resolve link prediction problem by training the utilized feed-forward neural network model by the aim of three nature-inspired algorithms including genetic algorithm, particle swarm optimization, and moth search (Yaghi, Faris, Aljarah, Ala'M, Heidari, & Mirjalili, 2020). Thus, while some features are generally hidden, another group of similarity metrics, named network structure-based similarity can be helpful. The structure-based similarity includes local-based/global-based, parameter-free/parameter-dependent, or node-dependent/path-dependent aspects. Mutlu et al. provide a diverse study on feature extraction techniques which deeply investigates similarity metrics (Mutlu & Oghaz, 2019). In addition to similarity-based approaches, link prediction can also be tackled by learning-based approaches (Nechaev, Corcoglioniti, & Giuliano, 2018), (Bastami, Mahabadi, & Taghizadeh, 2019) such as applying neural network-based candidate selection algorithm to harvest the vast amounts of data available on social networks (Gaudelet, et al., 2020). Mutlu et al. study maximum likelihood methods, probabilistic methods and graph representation learning approaches in link prediction problem (Mutlu & Oghaz, 2019). These methods provide the features on which machine learning models can learn and treat the problem as a binary classification task by the aim of some typical machine learning models such as classifiers and probabilistic models (Peng, BaoWen, YuRong, & XiaoYu, 2015).

Classification is a public method of learning based approaches. Some of the most powerful classification algorithms used in data mining techniques are classifiers of machine learning used to extract patterns from raw data (Wu, et al., 2008), including C4.5, K-means, Support vector machine (SVM), Apriori, AdaBoost, *k*-nearest neighbor classification (kNN), Bayesian classifiers, Classification and Regression Trees (CART) and so forth.

A comprehensive comparison of the precision of supervised and unsupervised classification algorithms on different local indices tested on Facebook dataset is provided by Raut (Raut, Khandelwal, & Vyas, 2020). The purpose of data mining is to choose the correct classifier method that includes high precision; some popular classifiers are decision trees, Bayesian networks and neural networks. The training dataset, which has predefined class labels, is given to a classification algorithm. Then, the classifier is tested by test dataset, where all class labels are hidden. In the case that the most items in the test dataset are correctly classified, the classifier can be assumed as an accurate one for current data. Fig. 1 depicts the mentioned process.

## 2.2 Related Works

There are some excellent surveys for the link prediction problem (Peng, BaoWen, YuRong, & XiaoYu, 2015), (Lü & Zhou, 2011), (Liben-Nowell & Kleinberg, 2007), (Al Hasan & Zaki, 2011), (Martínez, Berzal, & Cubero, 2017), (Martínez, Berzal, & Cubero, 2017), (Samad, Qadir, Nawaz, Islam, & Aleem, A comprehensive survey of link prediction techniques for social network, 2020), and (S & Sadasivam G, 2020). The link prediction algorithms are generally categorized into unsupervised and supervised methods. Neighborhood measures are basically used on unsupervised methods (Liben-Nowell & Kleinberg, 2007). Supervised methods (Lichtenwalter, Lussier, & Chawla, 2010) consider link prediction problems as classification ones. The latter are the state-of-the-art and generally provide more accurate results. Peng et al. (Peng, BaoWen, YuRong, & XiaoYu, 2015) discuss basic link prediction metrics including node-based metrics, topology-based metrics, social-theory-based metrics and learning-based methods to calculate the similarities of node pairs and provide a systematic category for link prediction methods. Meanwhile, some considered taxonomies of link prediction techniques (Martínez, Berzal, & Cubero, 2017), (Al Hasan & Zaki, 2011) introduces four groups of approaches including similarity, statistical, algorithmic, and preprocessing approaches which can be applied to different networks including simple, social and complex ones. Most of the studies on the link prediction literature mainly focus on the unsigned networks like Facebook and MySpace; some research including (Gu, Chen, Li, Liu, & Chen, 2019) and (Yuan, He, Guan, Zhou, & Li, 2019) study the link prediction approach in signed networks. A detailed and comprehensive study of the state-of-the-art of link prediction approaches and a study on the computational complexity analysis of most important techniques are provided in (Martínez, Berzal, & Cubero, 2017). Additionally, they perform an empirical study of the most important techniques applied to a set of networks with different properties. Our focus in this section is on the most general one unsupervised methods, called similarity-based approaches, which find the similarity between two nodes in the network structure by estimating the proximity between them. The mainstreaming classes of similarity-based link prediction techniques (Lü & Zhou, 2011) have three major categories namely local indices, global indices and Quasi-local indices (Mutlu & Oghaz, 2019). In some references, these categories are also named as node based and path based topological metrics (S & Sadasivam G, 2020). For a wide comparison and a deep understanding of the categories of indices, readers are encouraged to see (Martínez, Berzal, & Cubero, 2017), (S & Sadasivam G, 2020), (Samad, Qadir, Nawaz, Islam, & Aleem, A comprehensive survey of link prediction techniques for social network, 2020), (Dong, Ke, Wang, & Wu, 2011), (Haghani & Keyvanpour, 2019) and (Zhou, Lü, & Zhang, 2009). Merely using local or quasi-local indices leads to poor predictions in very sparse networks. Indirect connections in global indices raise the noise and computational complexity of the link prediction process. Having outlined the major link prediction methods, now the state-of-the-art methods to similarity-based link prediction approaches are reviewed.

Some of the local information based indices are studied in (Yuan, Ma, Zhang, Liu, & Shen, 2015) in order to find the impact of social network structures on the accuracy of similarity-based link prediction; e.g. Common Neighbors (CN), Adamic-Adar Index (AA), and Resource Allocation Index (RA). In addition to these three mentioned indices by Yuan et al., other local information similarity measures are introduced in (Martínez, Berzal, & Cubero, 2017), (Lü & Zhou, 2011), (S & Sadasivam G, 2020) and (Samad, Qadir, Nawaz, Islam, & Aleem, A comprehensive survey of link prediction techniques for social network, 2020) including Resource Allocation Based on Common Neighbor Interactions (RA-CNI), Preferential Attachment Index (PA), Jaccard Index (JA), Salton Index (SA), Sørensen Index (SO), Salton Cosine Similarity (SC), Hub Promoted Index (HPI), Hub Depressed Index (HDI), Local Leicht-Holme-Newman Index (LLHN), Individual Attraction Index (IA), Mutual Information (MI), Local Naive Bayes (LNB), CAR-Based Indices (CAR), Functional Similarity Weight (FSW), Local Interacting Score (LIT). Most link prediction techniques use pair of node as one unit and make decisions based on the commonality between them; by contrast, SAM index proposed by Samad et al. considers similarity of each node towards the other node (Samad, Qadir, & Nawaz, 2019). Moreover, Dong et al. (Dong, Ke, Wang, & Wu, 2011) propose two link prediction indices named Individual Attraction Index (IA), and its simple edition called Simple Individual Attraction Index (SIA) which take both the individual characteristics of common neighbors and the impact of the small network formed by the common neighbors. The proposed link prediction algorithm is based on not only node similarity, but also node's local information and relationships. A neighborhood-based similarity link prediction algorithm (Mumin, Shi, & Liu, 2019 ) combines a common neighbor score with the node degree of the neighbors; besides, it considers the impact of node degree in link prediction by ensuring fair allocation of resources to the various nodes.

In (Zhou, Lü, & Zhang, 2009) and (Dong, Ke, Wang, & Wu, 2011), nine well-known local-information-based similarity measures on some real networks are widely investigated and it is found that many links are mostly assigned the same scores in case that the information of the nearest neighbors is used. Although these algorithms, based on different similarity measures, are practically successful in dealing with specific networks, there is no comprehensive information about the dependence of algorithmic performance on network topology (Zhou, Lü, & Zhang, 2009). Besides, in (Liben-Nowell & Kleinberg, 2007), local information measures of similarity of nodes are computed, related to the network topology, indicating that the network topology contains latent information from which to infer future interactions. This research applies certain fairly subtle measures such as infinite sums over paths in the network which outperform more direct measures, such as shortest path distances and numbers of shared neighbors. In addition, (Pan, Li, Liu, & Liang, 2010) has presented an agglomerative community detection method based on node similarity, without requiring any prior knowledge about the number of communities in the networks; their proposed method starts with an arbitrary node to form new communities using only local information. The introduced link prediction method in (Aziz, Gul, Muhammad, & Uddin, 2020) aims at improving the accuracy of existing path-based methods by incorporating information about the nodes along local paths. Their proposed similarity index between two nodes $u$ and $v$ is computed by using not only the information of the nodes that are directly connected to $u$ an $v$, but we also incorporate the information about those nodes that lie on set of all possible paths of smaller length from $u$ to $v$. Martínez et al. (Martínez, Berzal, & Cubero, 2017), study some global similarity measures deeply including Negated Shortest Path (NSP), Katz index (KI), Leicht-Holme-Newman index (GLHN), Random walks (RW), Random walks with restart (RWR) and the like; for more detail, NSP computes the shortest path between two nodes using Dijkstra's algorithm, however it has a low prediction accuracy in comparison with local measures and other global methods (Liben-Nowell & Kleinberg, 2007). KI sums the influence of all possible paths between two pairs of nodes, incrementally penalizing paths by their length. GLHN index assigns a similarity proportional to the number of paths between nodes. To find RW, a neighbor of a node is randomly picked and move from the node to the neighbor; then, this process is repeated for each reached node and a Markov chain of randomly-selected nodes is created. If a probability is considered for returning to the starting node instead of moving forward, the model will be a RWR. Zhu et al. tackles a global link prediction on a sparse, bipartite multi-graph of an e-commerce dataset by the aim of GNN to find expressive embedding for link prediction. Transformer architecture is originally utilized for sequence transduction in language modeling and translation and use random and spectral embedding. The proposed approach of (Zhu, Li, & Somasundaram, 2019), named Temporal Graph Transformer (TGT), generates a product embedding for each customer which can be utilized to find likely items for each customer. Another related work based on topological structure (Zhao, Feng, Wang, Huang, Williams, & Fan, 2012) is topic modeling through analyzing the contents of social objects. In the first step, it clusters all social objects into different topics. Then, it divides the involved members in each social object into different topical clusters. Finally, it differentiates the strength of connections between members by detection of link-based community for each topical cluster. In another link prediction study (Berlusconi, Calderoni, Parolini, Verani, & Piccardi, 2016) which identifies missing links in a criminal network, links are classified as marginal on the basis of the topological analysis. The missing links have features contrary to marginal ones. In (Yang, Zhang, Zhu, & Tian, 2018), the network structure is investigated and the significant influence is determined according to the relations built through the paths between endpoints instead of the endpoint degree. Strong relations connecting the other endpoint through short paths, especially through common neighbors, can bring in more powerful influence; accordingly, a link prediction index SI is proposed, which deliberately models the significant influence by distinguishing the strong influence from the weak.

Aziz et al. propose a global and quasi-local extensions of some commonly used local similarity indices that applies the node information on local paths (Aziz, Gul, Uddin, & Gkoutos, 2020). Quasi-local indices provide a tradeoff between accuracy and computational time. A vector-based implementation of several local structural similarity indices with their global and quasi-local extensions is introduced in (Aziz, Gul, Uddin, & Gkoutos, 2020). Besides, a wide comparison of the local indices with their global and quasi-local extensions is provided and finally they conclude that the global and quasi-local indices usually have better performance although with much higher computational cost. Furthermore, in their experiments they find out that integration of global and quasi-local measures outperforms the Katz index. In (Zhao, Feng, Wang, Huang, Williams, & Fan, 2012), a topic oriented community detection approach is developed combining both social objects clustering and link analysis. Firstly, a subspace clustering algorithm is applied to categorize all social objects into topics. Then, the members of those social objects are assigned to topical clusters. Finally, to differentiate the strength of connections, a link analysis on each topical cluster is run in order to detect the topical communities. This method can identify communities from the perspective of both topics and the link structures which can predict people's attraction to communities and topics. An ensemble enabled link prediction approach is applied in (Duan, Ma, Aggarwal, Ma, & Huai, 2017). This approach decomposes traditional link prediction search space problems into sub-problems of smaller matrices solved with latent factor models, effectively proper to networks of modest size. The purpose of using ensemble approach is to reduce the sizes of sub-problems without sacrificing the prediction accuracy. Furthermore, three bagging methods that are developed in particular for link prediction are developed. Effective techniques are designed to reduce the network sizes of ensembles in the bagging without sacrificing the prediction accuracy. Although the article (Kossinets, 2006) pays particular attention to local criteria with less relevant to global ones, but in its bioinformatics studies, the clustering approach considers the membership of proteins in a group by combining local and global indices. The link prediction process runs on clusters using local similarity indices to find new links. Three major missing data mechanisms namely network boundary specification, survey non-response, and censoring by vertex degree are examined their impact on the scientific collaboration network from the Los Alamos E-print Archive and random bipartite graphs. Moreover, they present that social networks with multiple interaction contexts have certain surprising properties because of overlapping cliques.

Regarding (Haghani & Keyvanpour, 2019), Haghani classifies supervised link prediction approaches into two major groups including learning based methods and heuristic ones. With a learning look, different approaches are designed to employ learners with learning-based link prediction methods by use of similarity measures such as articles (Lü & Zhou, 2011), (Martínez, Berzal, & Cubero, 2017), (Peng, BaoWen, YuRong, & XiaoYu, 2015). Zhang applies a heuristic learning paradigm for link prediction in a way that learning is justified from local sub-graphs instead of entire networks (Zhang & Chen, 2018). Meanwhile, a γ-decaying theory is applied to unify a wide range of high-order heuristics and prove their approximability from local sub-graphs. Motivated by the theory, a link prediction framework based on GNN, named SEAL is produced. This framework simultaneously learns from local enclosing sub-graphs, embedding and attributes by the use of GNN. In (Sarkar, 2011), a link-prediction heuristic is tackled in a class of graph generation model with nodes associated with locations in a latent metric space and connections are more likely between closer nodes. a pre-processing step is considered in similarity-based prediction approaches. A sequence of formal results are provided including the relative importance of short paths versus long paths, node's degree plays bounds related role in its efficiency for link prediction process and the effects of increasing non-determinism in the link generation process on link prediction quality. The link prediction problem is formalized as distance estimation between nodes in a latent space. The introduced method has two stages; the first stage is a pre-processing phase and the second stage has the main processing actions. It is shown that by removing additional links in pre-processing stage, more accurate predictions can be made.

In fact, both the local information and global indices, associated with structural information of the nodes, are important for similarity-based link prediction. The existing studied approaches consider only one aspect but ignore the other one; the reviewed works either contain more than one local index, or have global indices. Although link prediction is not a new challenge of information technology, the studied methods cannot handle current development of the new perspectives appeared in social networks. To address this problem, we propose here an intelligent approach that brings these two categories of similarity-based link prediction methods together.

## 3. The Proposed Approach: Classifying Link Prediction

As outlined in related works, the reviewed works mostly focus on local parameters. The major advantages of applying local indices in link prediction problems are their low computation complexity and being suitable for large-scale and dynamic networks. Therefore, using local indices makes the prediction approaches faster than nonlocal; besides, these link prediction methods are highly more parallelizable (Martínez, Berzal, & Cubero, 2017); but due to the limited amount of information, the prediction accuracy of these approaches is widely low (Yuan, Ma, Zhang, Liu, & Shen, 2015). Thus, global measures are applied which have better accuracy with high computing complexity. In this study, the introduced classification based approach uses the combination of these two local and global measures in two phases; the first phase is a pre-processing and computation of measures, the local and global metrics. Finally, in second phase, the classification algorithm is executed to predict new links. Fig. 2 shows the proposed link prediction process. The first phase in the figure is offline and it runs one time. In this phase, the local and global parameters are computed by the aim of the existing data for each node. Then, our classification based link prediction model is constructed using this classification algorithm. The second phase is the usage of the extracted model; for example, in case that node *y* is studied to suggest a link to node *x* or not, firstly the related local and global similarity indices of these two nodes are computed. Then, the learning based model determines whether node *y* is recommended as a friend to node *x*.

## 3.1 Local and Global Indices

The proposed algorithm uses four parameters. The first three parameters are local and the fourth is a global one. Each of these indices specifies a part of the developed classification algorithm. The concept of *friend of friends* (*FoF*) is used for information computation. The local indices are computed as follows using *FoF* concept.

Let *v* refers to the node in the considered social network going to be analyzed. Suppose $C_v$ is the set of neighboring nodes of *v*. $D_v$, called as the *Clustering Coefficient* of node *v* (Silva, Tsang, Cavalcanti, & Tsang, 2010), shows the density of the neighboring nodes of *v*. It is defined as

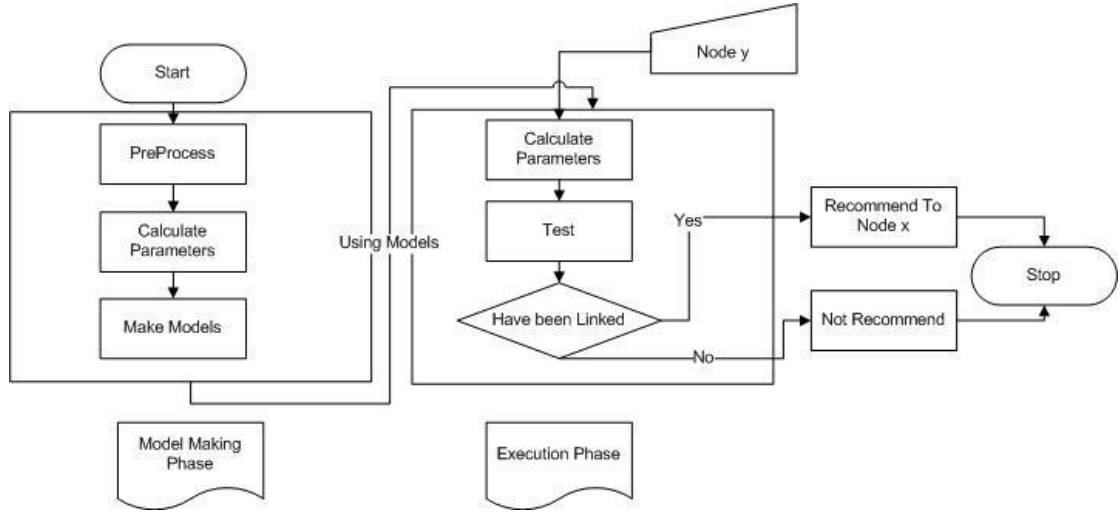

Fig. 2. The proposed link prediction process

$$D_v = \frac{\sum_{i \in C_v}(\sum_{j \in C_v}(M_{ij}))}{(|C_v|*(|C_v|-1))/2},$$
(Eq. 2)

where *M* is the adjacency matrix whose rows and columns represent the name of nodes. Its entries represent binary digits, where 1 means the existence of a link between two nodes, and 0 shows the absence of the link (Yaghi, Faris, Aljarah, Ala'M, Heidari, & Mirjalili, 2020). Therefore, the numerator in Eq. 2 computes the number of neighbors of the neighboring nodes of *v* ($C_v$); the denominator is the number of clique edges of

*v*. A clique is a subset of the vertices of an undirected graph, such that every two distinct vertices are adjacent.

The remainder of this section discusses the required indices.

The first index, $I_1$, refers to *FoF*, the number of common friends of nodes *u* and *v*, on the social network.

$$I_1 = |C_u \cap C_v|, \qquad \forall u, v \in V \tag{Eq. 3}$$

The second index, $I_2$, is the density of the first index, discussed in Eq. 3,

$$I_2 = D_{C_u \cap C_v}. \tag{Eq. 4}$$

Eq.4 specifies the level of adhesion (relations) between the common friends of nodes *u* and *v*. In social networks, if formed groups of common friends of nodes have many relations with one another the value of $I_2$ will increase. To compute $I_2$, firstly common friends of nodes *u* and *v* are determined, ($C_u \cap C_v$). Then, the density of their common friends is computed by using Eq. 2.

The third local index is the density of the formed groups by nodes *u* and *v*. The second index, $I_2$ computes the *clustering coefficient* (Eq. 2) of the intersection of neighboring nodes of the considered nodes, while $I_3$ computes the *clustering Coefficient* of the union of neighboring nodes of the considered nodes as follows,

$$I_3 = D_{C_u \cup C_v}. \tag{Eq. 5}$$

The last considered index is a global one (Symeonidis, Tiakas, & Manolopoulos, 2010), derived from Jaccard's benchmark which computes the similarity between two nodes as follows.

$$sim(u, v) = \frac{1}{\deg(u) + \deg(v) - 1}, \tag{Eq. 6}$$

where, deg(.) is the degree of the input node; sim(*u*,*v*) is in range of [0,1]; the more similar the nodes are, the more the value of Eq.6 will be close to 1 and vice versa.

The global index is computed using adjacency matrix as follows,

$$S[i, j] = sim(i, j) \qquad \forall i, j \in V. \tag{Eq. 7}$$

Eq.7 computes the similarity value of each two nodes of the graph; needless to mention that *S[i,j]* has the similarity values of the nodes with direct links to each other and the similarity value of nodes without any direct link is zero. This parameter seems to be a local one, but in the case that two nodes are not neighbors, global information is required. The similarity value of such nodes is computed using transitive similarity as follows,

$$sim(u, v) = \prod_{i=1}^{k} sim(iv_i, iv_{i+1}). \tag{Eq. 8}$$

All the intermediate nodes in the shortest path from *v* to *j* passes through are considered as ***iv*** = {$iv_1, iv_2, \ldots, iv_k$}, where $iv_1 = v$ and $iv_{k+1} = j$. Regarding Eq.8, in the case that two nodes have not direct link, the similarity value is computed by multiplying the similarity values of the shortest path between the two nodes. Dijkstra's algorithm is used to find the shortest path between them. As in our case, as the edges have the same costs the shortest path is the one with the least number of edges.

After calculating the indices, the model is formulated as follows.

## 3.2 Designing the Proposed Model

After computing the indices, classification algorithm is applied. As mentioned in 2.1.1, different algorithms are used for classification. In this study, AdaBoost algorithm is used as one of the best out-of-the-box classifiers (Martínez, Berzal, & Cubero, 2017); it is one of the most important ensembles learning methods dealing with ones that employ multiple learners to solve a problem. It has a great simplicity with solid theoretical foundation and accurate prediction (Kégl, 2013). This algorithm aims to convert a set of weak classifiers into a stronger one. Unlike most issues in the field of machine learning that suffer from many dimensions, AdaBoost algorithm selects only features that increase the model prediction capability, thereby reduce the dimensionality and, since it does not calculate irrelevant features; this characteristic potentially improves time complexity of the classifier (Kégl, 2013).

Classifiers might individually be weak so combining them with learning algorithms can converge to a stronger model; AdaBoost can be applied to any of such learning algorithms. Here, decision tree as a weak classifier is used with AdaBoost. The goal of AdaBoost is firstly to choose a proper training set for each learning iteration of forming the decision tree and secondly, to determine the probability of outputs of the decision tree. To determine the probability, the AdaBoost algorithm initially assigns equal weights to all members. A base learner is generated using the training set; it is worth mentioning that the distribution of weights of training set members, at step $t$, is demonstrated by $D_t$. The base learner is tested the weights and improves them to $D_{t+1}$ at learning iteration $t+1$ based on their precision; so the new weights will be made up better. It is assigned as the training set of the next decision tree, and hopefully leads to a much better performance due to improving the weights. Then, the decision tree is generated and the process is repeated for $T$ rounds. Eventually, the final output is a linear combination of all weak classifiers outputs, and the ultimate decision is simply made based on the sign of the summation result; weighted majority voting of $T$ weak learners determines the final model.

The phases of the proposed classification model are discussed as follows.

Suppose a data set containing $n$ samples and $m$ labels, in form of $(x_i, y_i)$, where $x_i \in X$ is sample $i$ in our considered $n$ samples and $y_i \in Y$ is the label of a class to which $x_i$ belongs. Initially, in step $1$, weight of each sample is assigned $D_1(x_i, y_i) = \frac{1}{n}$. The given weak learner, decision tree, is applied to find a weak hypothesis named $h_t: X \rightarrow Y$ with a low weighted classification error $\varepsilon_t = Pr_i \sim D_t[h_t(x_i) \neq y_i]$. Weight of the weak classifier is $\theta_t = \frac{1}{2} ln\left(\frac{1-\varepsilon_t}{\varepsilon_t}\right)$; finally, the probabilities of data samples are updated using Eq.9.

$$D_{t+1}(x_i, y_i) = \frac{D_t(x_i, y_i) e^{[-\theta_t y_i h_t(x_i)]}}{Z_t},$$ (Eq. 9)

where $Z_t$ is sum of all the weights. As the summation of these probabilities should be 1, the weights are normalized by dividing by $Z_t$. Function $e^x$ returns a fraction for negative values of $x$, and a value greater than one for positive values. Thus, the updated weight is changed depending on the final sign of the term '-$\theta_t y_i h_t(x_i)$'. Previously mentioned, $y_i$ is the actual label of $x_i$ and $h_t(x_i)$ is the predicted label of $x_i$. In the case that $y = \{-1, 1\}$, if they both are the same, the sign of term '$-\theta_t y_i h_t(x_i)$' does not change, while misclassification changes the sign which leads to a decrease of the updated weight.

## 4. Evaluation

To evaluate our proposed algorithm in this study WEKA project is applied; WEKA provides a comprehensive collection of machine learning algorithms and data preprocessing tools (Hall, Frank, Holmes, Pfahringer, Reutemann, & Witten, 2009). This workbench has algorithms of regression, classification, clustering, association rule mining and attribute selection. Data visualization facilities and many preprocessing tools are enabled by data exploration. Combining these features with statistical evaluation of learning schemes and learning results visualization provides data mining process which widely benefits our proposed approach.

The inputs are inserted in form of a relational table in ARFF. The output is a list of friend suggestions presented to a user; then the user may accept or reject the suggestions as friends. To evaluate such a problem of friend suggestion, it is imperative to have a real social network. In (Ahmed, 2015), (Cai, 2015), (Sengupta, 2015), (Symeonidis, Tiakas, & Manolopoulos, 2010) Facebook social network data and a number of other social networks are used. Facebook dataset is applied in our experiments as an input for global social network. This dataset identifies users' communications on Facebook over a period of time (Ahmed, 2015), (Cai, 2015), (Sengupta, 2015), (Symeonidis, Tiakas, & Manolopoulos, 2010). Additionally, our second applied global social network dataset is HepTh (High Energy Physics Theory), used by a wide range of researchers (Samad, Qadir, & Nawaz, 2019), (Dong, Ke, Wang, & Wu, 2011). Finally, the Brightkite dataset (Cho, 2011), (Hong, 2012), (Lü L. M., 2012) is considered as our third applied dataset. Brightkite is a location-based social network with 58228 nodes and 214078 edges. Although the original grid graph is assumed to be undirected it is supposed as a directed one in the dataset and each edge in this network contains an opposite edge as well. The applied datasets are extracted from the Stanford University dataset.

### 4.1 Preprocessing Phase

In this phase, after examining the input data, only the two-way communications are remained. In other words, one-way communications are initially emitted. Thus, after the pre-processing phase, the number of nodes and edges are astonishingly decreased. Table 1 lists the applied global and local datasets with the number of nodes and edges that exist in each dataset.

Table1. Input dataset

| dataset | Number of nodes/edges before pre-processing | Number of nodes/edges after pre-processing |
|---|---|---|
| Facebook | 4039 / 88234 | 4039 / 88234 |
| HepTh | 9877/25998 | 9877/25998 |
| Brightkite | 58228 / 428156 | 55723 / 214078 |

As presented in Table 1, the number of nodes and edges in the global datasets is the same before and after the pre-processing phase. The reason is that in these datasets all edges are bidirectional. So after the pre-processing, there is no change in the number of nodes and edges. The algorithm is run after this phase, to compute the introduced indices.

## 4.2 Performance Evaluation

Each user can run the algorithm and receives a list of suggestions as the output of the algorithm. The performance of the algorithm is determined based on the user's approval or rejection. However, in this study, due to the lack of access to the real social network, cross-validation method is applied to evaluate the performance of the proposed approach (Chen, 2014), (Sherkat, 2015), (Kohavi, 1995). The dataset is randomly divided into 10 parts; nine parts are used as training-set data and one part as the test-set data.

Common parameters for evaluation of the prediction algorithms are precision and recall computed using True Positive (TP) and False Positive (FP) criteria which indicate the accuracy and the completeness of the algorithm, respectively. They are computed as follows.

$$Precision = \frac{TP}{TP+FP}, \quad (Eq.10)$$

precision is a metric that quantifies the number of correct positive predictions made. TP (True Positive) depicts the number of predictions that truly predicts the sample belongs to the positive class and FP (False Positive) depicts the number of predictions assumed the sample to belong to the positive class while it is in the negative class.

$$Recall = \frac{TP}{TP+FN}, \quad (Eq.11)$$

recall is a metric that quantifies the number of correct positive predictions made out of all positive predictions that could have been made. FN (False Negative) depicts the number of predictions assumed the sample to belong to the negative class while the sample is in the positive class.

To evaluate the binary clusters based on both of the precision and the recall criteria, the fitness parameter is used. It is also known as harmonic mean of recall and precision (Samad, Qadir, Nawaz, Islam, & Aleem, 2020). As can be observed from Eq. 12, the fitness parameter computes the area under the curve.

$$Fitness = \frac{2 \times Recall \times Precision}{Recall + Precision}. \quad (Eq.12)$$

## 4.3 Proposed Algorithm Evaluation

In our experiments, Facebook, HepTh and Brightkite datasets are applied in order to evaluate the performance of the proposed approach. The values of the aforementioned parameters including precision, recall and fitness are evaluated in the experiments as follows (Fig. 3 to Fig. 5).

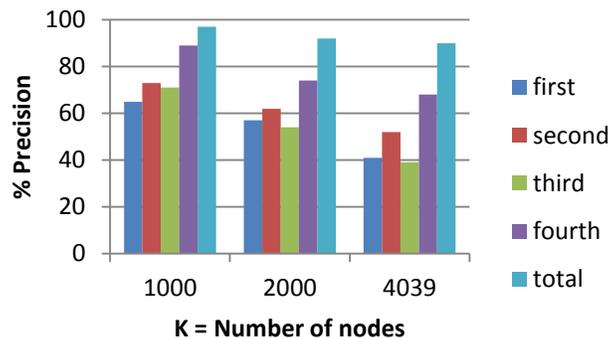

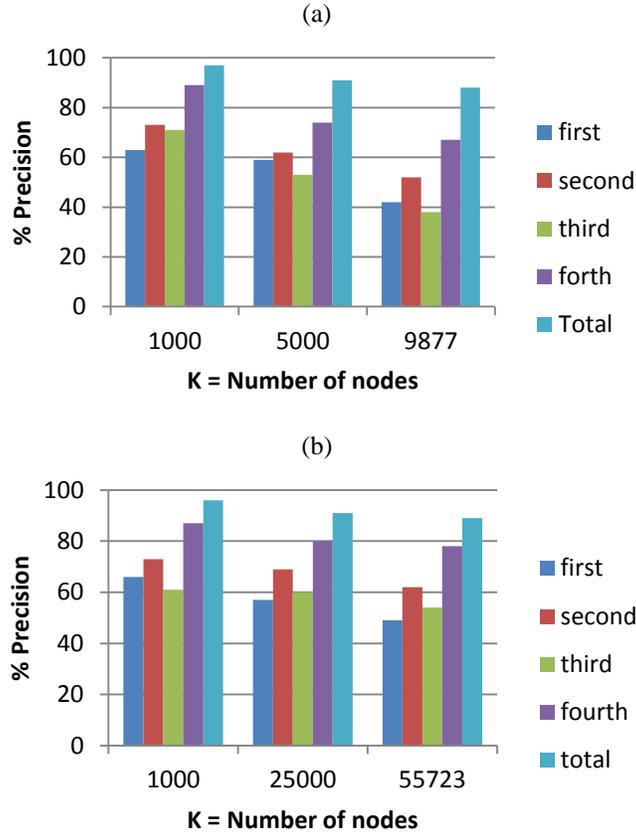

(c)

Fig. 3. The precision results for (a) Facebook, (b) HepTh, (c) Brightkite

The experiments are run with different number of nodes. For example, when *k* is 1000, it means that only 1000 nodes in the dataset are selected. Some same patterns can conveniently be observed from all of the graphs (Fig. 3 to Fig. 5) for all datasets. The bars in diagrams signify considered indices introduced in subsection 3.1 including three local parameters, one global parameter and total parameter as the application of all parameters together namely first, second, third, fourth and total, in the order mentioned. Fig. 3 shows the precision values for the Facebook dataset (a), the HepTh dataset (b) and the Brightkite dataset (c). As this figure depicts, the more the number of nodes, the lower the precision; undoubtedly, greater number of nodes means more communications among the nodes. Additionally, higher values of indices (such as the larger number of common neighbors among nodes for $I_1$) lead to increasing the complexity of computations in a way that the precision decreases. In Fig. 3, lower precision can obviously be observed when local parameters are used. The precision of predictions increases when the fourth parameter is applied. Needless to say that it is because the fourth parameter basically takes into account the global information of the network. Eventually, the precision hits the peak when both local and global parameters are used together. The discussed trends are apparent in the results of the evaluation of all datasets depicted in Fig. 3. Results from Facebook are shown in Fig. 3(a). It shows that the precision of predictions when the proposed approach is used is much higher than using other parameters (approximately 90 to 97), while the usage of the global parameter results around 68 to 89. Besides the local parameters have not considerable precision in comparison with the fourth and total parameter. The same pattern can be observed in Fig. 3(b) and Fig.

3(c) for the HepTh and the Brightkite, respectively. Moreover, Fig. 3 shows that the integration of local and global parameters in link prediction process on Facebook dataset has the best precision results.

Similarly, Fig. 4 shows the recall values for the Facebook dataset (a), the HepTh dataset (b) and the Brightkite dataset (c). All charts have the same trends that the recall increases when the number of nodes grows. Like precision, the higher values of recall show better performance of the executed prediction. Thus, the prediction has the least recall in case of using the first parameter. Then, the third parameter has the next least value. After that the second parameter comes to the surface. The recall of the prediction is better when global parameter (the fourth parameter) is applied. Finally, as can be observed from Fig. 4, when both local and global indices are integrated (total parameter) the recall values are the best in all datasets. Fig. 4(a) addresses the prediction recalls of 1000 to 4039 nodes from Facebook dataset. Again total parameter outperforms other measures on all experiments ranging from 90 to 97 for 4039 to 1000 number of nodes. The maximum recall achieved by total is 97 on dataset Facebook when all nodes are considered. It is worth mentioning that the best values of recall on two other datasets namely HepTh and Brightkite are high as well (95 and 94), respectively. Overall, the recall values of predictions have nearly the same pattern for all datasets including HepTh and Brightkite. It can be conveniently observed from Fig. 4 that integration of local and global parameters in predicting new links has the best recall in Facebook dataset.

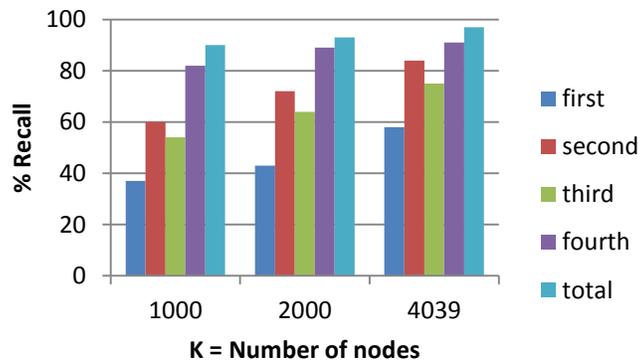

(a)

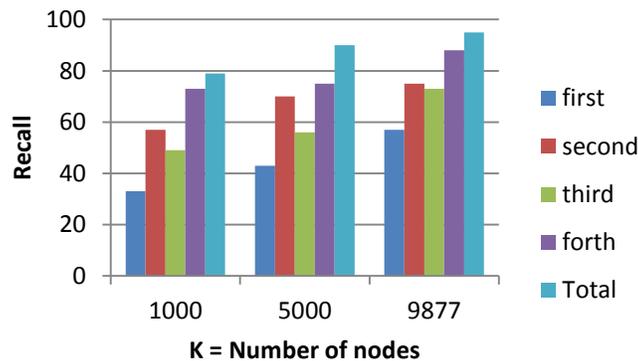

(b)

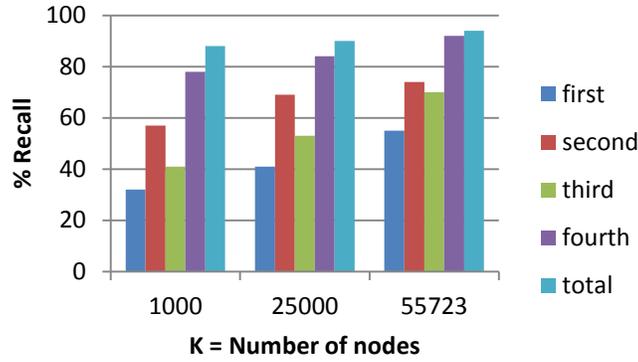

(c)

Fig. 4. The recall results for (a) Facebook, (b) HepTh, (c) Brightkite

As stated in section 4, the precision and recall parameters are not sufficient and the fitness is also applied in this study to provide a way to combine both the precision and recall parameters into a single measure that captures both properties. Fig. 5 shows the results of the fitness parameter (introduced in Eq. 12) for Facebook dataset (a), the HepTh dataset (b) and Brightkite dataset (c). In all datasets, the values of fitness are the highest when both local and global parameters are applied. Besides, in all diagrams as the number of nodes increases, the fitness increases as well.

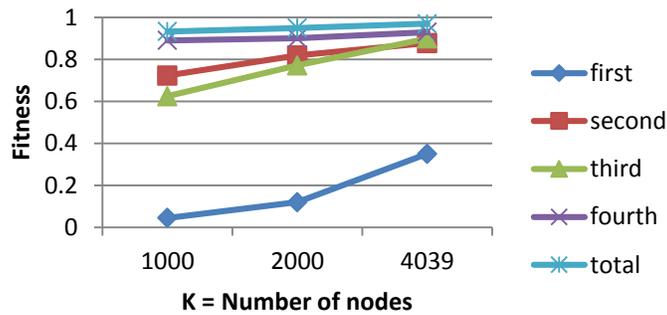

(a)

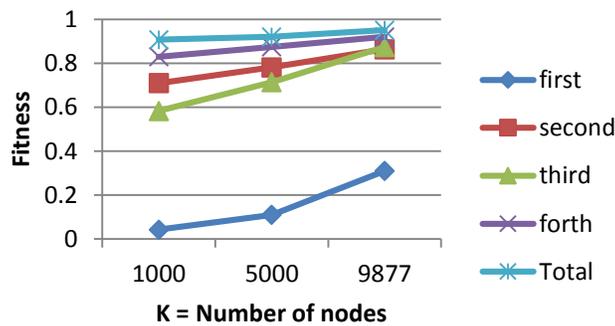

(b)

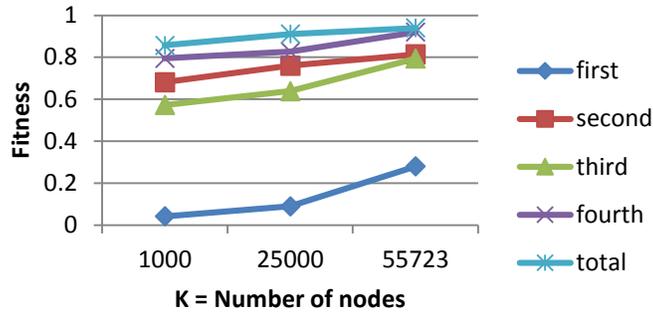

(c)

Fig. 5. The fitness results for (a) Facebook, (b) HepTh, (c) Brightkite

Regarding Fig. 5(a), the fitness of performed experiment on Facebook depict that the results of the local parameters are much lower than the results of the global parameter. Moreover, using both local and global parameters shows the best values of fitness on Facebook, for all values of *K*. The same trend is obviously observable on the other datasets (HepTh and Brightkite). Additionally, Table 2 shows the fitness results in detail for all of the datasets including Facebook, HepTh and Brightkite.

Table. 2 The fitness results for different datasets while applying different indices

| Facebook nodes | HepTh nodes | Brightkite nodes | Applied index | Facebook Fitness | HepTh Fitness | Brightkite Fitness |
| --- | --- | --- | --- | --- | --- | --- |
| 1000 | 1000 | 1000 | First | 0.045 | 0.043 | 0.042 |
| | | | Second | 0.723 | 0.71 | 0.681 |
| | | | Third | 0.625 | 0.583 | 0.573 |
| | | | Fourth | 0.891 | 0.83 | 0.795 |
| | | | Total | 0.933 | 0.907 | 0.857 |
| 2000 | 5000 | 25000 | First | 0.12 | 0.11 | 0.09 |
| | | | Second | 0.891 | 0.781 | 0.76 |
| | | | Third | 0.772 | 0.714 | 0.639 |
| | | | Fourth | 0.901 | 0.875 | 0.827 |
| | | | Total | 0.95 | 0.92 | 0.911 |
| 4039 | 9877 | 55723 | First | 0.35 | 0.311 | 0.28 |
| | | | Second | 0.878 | 0.864 | 0.814 |
| | | | Third | 0.9 | 0.874 | 0.794 |
| | | | Fourth | 0.93 | 0.92 | 0.92 |
| | | | Total | 0.9706 | 0.951 | 0.94 |

Table. 3 The prediction accuracy of the proposed (Total) and the alternate methods, measured by Fitness

| Facebook nodes | HepTh nodes | Brightkite nodes | Applied index | Facebook Fitness | HepTh Fitness | Brightkite Fitness |
|---|---|---|---|---|---|---|
| 1000 | 1000 | 1000 | SAM | 0.895 | 0.862 | 0.82 |
| | | | CN2D | **0.936** | 0.88 | **0.853** |
| | | | PSI | 0.921 | 0.894 | **0.85** |
| | | | SI | 0.866 | 0.735 | 0.791 |
| | | | Total | **0.933** | **0.907** | **0.857** |
| 2000 | 5000 | 25000 | SAM | 0.92 | 0.892 | 0.868 |
| | | | CN2D | **0.96** | 0.897 | 0.907 |
| | | | PSI | 0.946 | 0.89 | 0.866 |
| | | | SI | 0.879 | 0.784 | 0.82 |
| | | | Total | **0.95** | **0.92** | **0.911** |
| 4039 | 9877 | 55723 | SAM | 0.933 | 0.923 | 0.917 |
| | | | CN2D | **0.993** | 0.94 | **0.948** |
| | | | PSI | 0.96 | 0.937 | 0.922 |
| | | | SI | 0.892 | 0.79 | 0.836 |
| | | | Total | **0.971** | **0.951** | **0.94** |

It is worth mentioning that the required time for computing the global indices is high. Thus, execution time of the performed experiments sometimes approaches to hours.

In order to evaluate the proposed approach compared to other techniques, the precision, recall and fitness of our link prediction method is compared with the one designed in (Samad, Qadir, & Nawaz, 2019). The experiments run on different datasets including Facebook, HepTh and Brightkite. In the light of the aforementioned experiments results discussed in the preceding paragraphs, we chose total measure to compare with SAM (Samad, Qadir, & Nawaz, 2019) to evaluate the performance of our proposed link prediction algorithm. SAM is a similarity measure that treats both nodes individually to compute similarity. This measure considers similarity of $u$ towards $v$ ($Sam(u_v) = \frac{|\tau(u) \cap \tau(v)|}{|\tau(u)|}$) as well as similarity of $v$ towards $u$ ($Sam(v_u) = \frac{|\tau(u) \cap \tau(v)|}{|\tau(v)|}$). The similarity of nodes $u$ and $v$ is computed by the average of $Sam(u_v)$ and $Sam(v_u)$. The proposed link prediction approach in (Samad, Qadir, & Nawaz, 2019) is applied in the experiments to compare with our approach. Additionally, in (Mumin, Shi, & Liu, 2019 ) the introduced index named CN2D calculates the similarity between two nodes using their common neighbor scores and resources distributed to the common neighbors based on their node degrees. The higher the similarity between them, the more likely that they would connect in the future. Another similarity index used in our performance evaluation is PSI (Aziz, Gul, Uddin, & Gkoutos, 2020) that computes similarity between two nodes $u$ and $v$; PSI uses damping factor by assigning the weights in such a way that the nodes at a distance greater than 1 get lower weights, and l is the length of the longest path that is considered in computation of PSI($u$, $v$). Finally, SI index (Yang, Zhang, Zhu, & Tian, 2018) which computes the significant influence of endpoints for link prediction. it is worth mentioning that two-hop paths bring in the significant influence while paths with three or more hops produce the weak influence in transferring resource. Therefore, Yang et al. separated two-hop paths by the common neighbors from the long paths. Besides, to enhance the prediction performance, an adjustable parameter $\alpha$ is set to penalize the weak influence derived from long paths.

The results of these comparisons are depicted in Table 3. As is represented in Table 3, it is noticed that our link prediction approach named as Total in illustrations, performs better than most of the other methods in terms of the fitness score. This validates the assertion that when both local and global indices are exploited using datamining techniques it can boost the prediction performance. It is also observed that our approach continuously achieves better results than SAM, PSI, and SI indices in all the experimented datasets. In particular, for Facebook dataset, CN2D overshadows the prediction fitness of other approaches. Moreover, it is very competitive when compared with BrightKite results.

## 5. Conclusion and future works

Regard to the exponential growth of communication between peers of a social network, predicting the future behaviors of members make the analysis and improvements of such networks more productive and effective. Different methods are provided in current studies including similarity based, learning based, and probabilistic statistical approaches. In this research, a combination of approaches is applied; firstly, some new similarity measures including local and global indices are extracted. Then, learning based link prediction approaches, AdaBoost classifier algorithm with the aims of decision tree is used in link prediction problem. Both local and global indices are used to improve the performance of the algorithm. The experiments present well performance of the proposed method in comparison with using just one link prediction approach. Moreover, our proposed link prediction approach has a better performance in comparison with recent studied works.

Our future work will mainly emphasize on link prediction problem in signed social networks. Many effective link prediction models are used to achieve positive and negative links prediction results. We will try to generate such information to increase the predicted link information.